\documentclass[sigconf]{acmart}
\usepackage{multirow}
\usepackage{enumitem}
\usepackage{subcaption}

\AtBeginDocument{%
  \providecommand\BibTeX{{%
    \normalfont B\kern-0.5em{\scshape i\kern-0.25em b}\kern-0.8em\TeX}}}

\begin{document}
%% 8pages main content
% \title{The Name of the Title is Hope}

\title{Asymptotically Unbiased Estimation for Delayed Feedback Modeling via Label Correction}

% \author{Brooke Aster, David Mehldau}
% \email{dave,judy,steve@university.edu}
% \email{firstname.lastname@phillips.org}

\author{Yu Chen$^{ \dagger}$, Jiaqi Jin$^{ \dagger}$, Hui Zhao$^{\ddagger}$, Pengjie Wang,  Guojun Liu, Jian Xu and Bo Zheng$^{\scriptscriptstyle *}$}
\thanks{$^{ \dagger}$Co-first authorship. 
$^{\ddagger}$ This author is the one who gives a lot of guidance in the work
$^{\scriptscriptstyle *}$Corresponding author.}
\affiliation{
  \institution{Alibaba Group} 
%   \city{Hangzhou}
  \country{}
}
\email{{shuyuan.cy,jinjiaqi.jjq,shuqian.zh,pengjie.wpj,guojun.liugj,xiyu.xj,bozheng}@alibaba-inc.com}

% \setcopyright{acmcopyright}
% \copyrightyear{2018}
% \acmYear{2018}
% \acmDOI{10.1145/1122445.1122456}

% %% These commands are for a PROCEEDINGS abstract or paper.
% \acmConference[Woodstock '18]{Woodstock '18: ACM Symposium on Neural
%   Gaze Detection}{June 03--05, 2018}{Woodstock, NY}
% \acmBooktitle{Woodstock '18: ACM Symposium on Neural Gaze Detection,
%   June 03--05, 2018, Woodstock, NY}
% \acmPrice{15.00}
% \acmISBN{978-1-4503-XXXX-X/18/06}

\copyrightyear{2022}
\acmYear{2022}
\setcopyright{acmcopyright}
\acmConference[WWW '22]{Proceedings of the ACM Web Conference 2022}{April 25--29, 2022}{Virtual Event, Lyon, France}
\acmBooktitle{Proceedings of the ACM Web Conference 2022 (WWW '22), April 25--29, 2022, Virtual Event, Lyon, France}
\acmPrice{15.00}
\acmDOI{10.1145/3485447.3511965}
\acmISBN{978-1-4503-9096-5/22/04}

% \author{Yu Chen}
% \authornote{both authors contributed equally to this research.}
% \affiliation{%
%   \institution{Alibaba Group}
%   \city{Hangzhou}
%   \country{China}
% }
% \email{shuyuan.cy@alibaba-inc.com}

% \author{Jiaqi Jin}
% \affiliation{%
%   \institution{Alibaba Group}
%   \city{Hangzhou}
% \email{jinjiaqi.jjq@alibaba-inc.com}

% \author{Hui Zhao}
% \affiliation{%
%   \institution{Alibaba Group}
%   \city{Hangzhou}
% \email{shuqian.zh@alibaba-inc.com}

% \author{Pengjie Wang}
% \affiliation{%
%   \institution{Alibaba Group}
%   \city{Hangzhou}
% \email{pengjie.wpj@alibaba-inc.com}

% \author{Guojun Liu}
% \affiliation{%
%   \institution{Alibaba Group}
%   \city{Hangzhou}
% \email{guojun.liugj@alibaba-inc.com}

% \author{Jian Xu}
% \affiliation{%
%   \institution{Alibaba Group}
%   \city{Hangzhou}
% \email{xiyu.xj@alibaba-inc.com}

% \author{Bo Zheng}
% \authornote{Corresponding author: Bo Zheng.}
% \affiliation{%
%   \institution{Alibaba Group}
%   \city{Hangzhou}
% \email{bozheng@alibaba-inc.com}

\renewcommand{\shortauthors}{Yu Chen, et al.}

% \shortauthors{AA,BB,CC}
\begin{abstract}
Alleviating the delayed feedback problem is of crucial importance for the conversion rate(CVR) prediction in online advertising. Previous delayed feedback modeling methods using an observation window to balance the trade-off between waiting for accurate labels and consuming fresh feedback. Moreover, to estimate CVR upon the freshly observed but biased distribution with fake negatives, the importance sampling is widely used to reduce the distribution bias. While effective, we argue that previous approaches falsely treat fake negative samples as real negative during the importance weighting and have not fully utilized the observed positive samples, leading to suboptimal performance. 

In this work, we propose a new method, \textbf{DE}layed \textbf{F}eedback modeling with \textbf{U}nbia\textbf{S}ed \textbf{E}stimation, (DEFUSE), which aim to respectively correct the importance weights of the immediate positive, the fake negative, the real negative, and the delay positive samples at finer granularity. Specifically, we propose a two-step optimization approach that first infers the probability of fake negatives among observed negatives before applying importance sampling. To fully exploit the ground-truth immediate positives from the observed distribution, we further develop a bi-distribution modeling framework to jointly model the unbiased immediate positives and the biased delay conversions. Experimental results on both public and our industrial datasets validate the superiority of DEFUSE. Codes are available at \url{https://github.com/ychen216/DEFUSE.git}.
\end{abstract}






\maketitle
\begin{figure}[tbp]
    \centering
    \includegraphics[width=\linewidth]{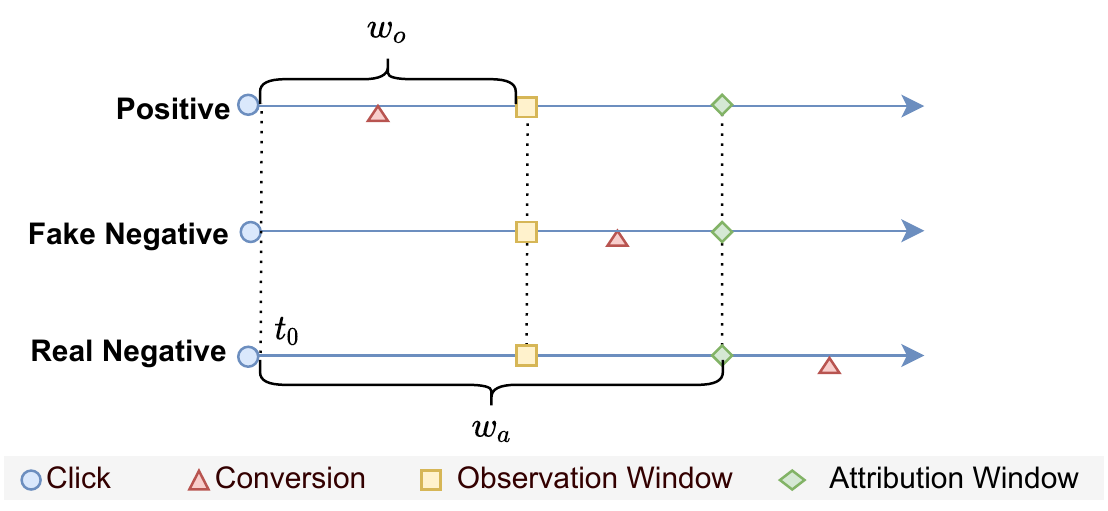}
    \caption{An illustration of different label types. The observation window $w_0$ denotes the least time interval between the click time and the streaming training time; while the attribution window $w_a$ determine the actual label. }
    \label{fig:delay_click}
\end{figure}
\section{Introduction}
Online advertising has become the primary business model for intelligent e-commerce, which helps advertisers target potential customers~\cite{evans2009online,goldfarb2011online,lu2017practical}. Generally, cost per action (CPA) and cost per click (CPC) are two widely used payment options, which directly influence both revenue of the platform and the return-on-investment (ROI) of all advertisers. 
As a fundamental part of both kinds of price bidding, conversion rate(CVR) prediction, which focuses on ROI-oriented optimization, always keeps an irreplaceable component to ensure a healthy advertising platform~\cite{lee2012estimating}.

As a widely used training framework, streaming learning, which continuously finetunes the model according to real-time feedback, has shown promising performance in click-through rate(CTR) prediction tasks~\cite{DBLP:journals/tsc/SongTS16,DBLP:conf/cikm/MoonLCLZC10,DBLP:conf/ijcai/SahooPLH18,DBLP:journals/corr/LiuXXZ17}. 
However, as shown in Table~\ref{tab:delay_distribution}, it is non-trivial to achieve better results via streaming learning due to the pervasively delayed and long-tailed conversion feedback for CVR prediction.
% $w_a$ which depends on different business scenarios
More specifically, as illustrated in Figure~\ref{fig:delay_click}, a click that happened at time $t_0$ needs to wait for a sufficiently long attribution window $w_a$ to determine its actual label --- only samples convert before $t_0 + w_a$ are labeled as positive. Typically, the setting for $w_a$ ranges from one day to several weeks for different business scenarios. The issue, even for an attribution window that as short as one day is still too long to ensure sample freshness, which remains a major obstacle for achieving effective streaming CVR prediction.

%The dilemma is, a relatively smaller $w_a$ provides the ability to use more fresh samples at the expense of using more inaccurate samples, 

To solve this challenge, existing efforts focus on introducing a much shorter observation window $w_o$, e.g., 30min, allowing clicks with observed labels to be collected and distributed to the training pipeline right after $t_0 + w_o$. Optimizing $w_o$ provides the ability to balance the trade-off between utilizing more fresh samples and accepting less accurate labels.
This greatly improves sample freshness, with acceptable coverage of conversions within the observation window, at the cost of temporarily marking feedback with long delay as \textit{fake negative}. Thus, current works mainly focus on making CVR estimations upon the freshly observed but biased distribution with fake negatives. 

Since it is hard to achieve unbiased estimation by using standard binary classification loss, e.g. cross-entropy, current efforts implement various auxiliary tasks to model conversion delay, so as to alleviate the bias caused by the fake negatives. 
Early methods~\cite{DBLP:conf/kdd/Chapelle14,DBLP:journals/corr/abs-1802-00255} attempts to address the delayed feedback problem by jointly optimizing CVR prediction with a delay model that predicts the delay time $d$ from an assumed delay distribution. 
However, these approaches are directly trained on the biased observed distribution and have not fully utilized the rare and sparse delayed positive feedback. 
Having realized such drawbacks, recent studies mainly focus on reusing the delayed conversions as positive samples upon conversion. Various sample duplicating mechanisms have been designed to fully exploit each conversion. 
For instance, FNC/FNW~\cite{DBLP:conf/recsys/KtenaTTMDHYS19} set $w_o=0$ and re-sent all positive samples when conversion arrives.
ES-DFM~\cite{DBLP:conf/aaai/YangLHZZZT21} only duplicate delay positive samples which previously have been incorrectly labeled as fake negatives; While DEFER~\cite{DBLP:conf/kdd/GuSFZZ21} reuse all samples with actual label after completing the label attribution to maintain equal feature distribution and to utilize the real negative samples. Moreover, to bridge the distribution bias, importance sampling~\cite{bishop2006pattern} is adopted to correct the disparity between the ground-truth and the observed but biased distribution.

\begin{table}[t]
    \centering
    \caption{The distribution for delay conversion on Taobao and Criteo datasets, where Inc and Acc respectively denote the incremental and accumulate proportion.}
    \begin{tabular}{c|cc|cc}
        \toprule
         Datasets & \multicolumn{2}{c|}{Taobao Dataset} & \multicolumn{2}{c}{Criteo} \\
         \midrule
         % \cline{3-8}
        delay interval & Inc (\%) & Acc (\%) & Inc (\%) & Acc (\%) \\
        \midrule
        \midrule
        <30min & 61 & 61 & 42 & 42\\
        30min - 12hour & 13 & 74 & 14 & 56\\
        12hour - 1day & 4 & 78 & 5 & 61 \\
        1day - 3day & 7 & 85 & 10 & 71 \\
        3day - 7day & 6 & 91 & 10 & 81 \\
        7day - 30day & 9 & 100 & 19 & 100 \\
        \bottomrule
    \end{tabular}
    \label{tab:delay_distribution}
\end{table}
Despite effectiveness, we argue that current methods still have some limitations. Firstly, they mainly focus on designing appropriate training pipelines to reduce the bias in feature space and only weight the loss of observed positives and negatives through importance sampling. The issue is, observed negatives may potentially be fake negatives, and these methods falsely treat them as real negatives, leading to sub-optimal performance. Second, observed positives can be further divided into \textit{immediate positives}(IP) and \textit{delay positives}(DP), which implies two potential improvements: (1) Intuitively, IPs and DPs contribute differently to the CVR model due to duplication. (2) By excluding DPs, an unbiased estimation of IP prediction can be established directly based on the observed dataset consistent with the actual distribution of IPs.

% % p5：引出本文的解决方案：1）IS修正；2）窗口内外建模减弱delay建模的偏差带来的影响
% % 等method部分完成后完善
In this paper, We propose \textbf{DE}layed \textbf{F}eedback modeling with \textbf{U}nbiased \textbf{E}stimation~(DEFUSE) for streaming CVR prediction, which investigates the influence of fake negatives and makes full use of DPs on importance sampling. Distinct from previous methods only modelling observed positives and negatives, we formally recognize the samples into four types, namely immediate positives(IP), fake negatives(FN), real negatives(RN), and delay positives(DP). Since FNs are adopted in the observed negatives, we propose a two-step optimization, which firstly infers the probability of observed negatives being fake negatives before performing unbiased CVR prediction via importance sampling on each of the four types of samples. Moreover, we design a bi-distribution framework to make full use of the immediate positives.
Comprehensive experiments show DEFUSE achieves better performance than the state-of-the-art methods on both public and industrial datasets.

% p6 contribution：1）新的IS修正思路，不同于以往专注于回补；2）充分利用了服从真实分布的窗口内样本，并提出了一种聚合机制；3）在公开数据和生产数据实验验证了有效性
Our main contributions can be summarized as follows:
\begin{itemize}[leftmargin=*]
    \item We emphasize the importance of dividing observed samples in a more granular manner, which is crucial for accurate importance sampling modeling.
    \item We proposed an unbiased importance sampling method, DEFUSE, with two-step optimization to address the delayed feedback issue. Moreover, we implement a bi-distribution modeling framework to fully exploit immediate positives during streaming learning.
    \item We conduct extensive experiments on both public and industrial datasets to demonstrate the state-of-the-art performance of our DEFUSE.
\end{itemize}

% 需要确认的一些key words:
% 模型名称，窗口内外建模，重要性采样修正，我们的数据集名称，联合分布优化，EM优化。

\section{Related Work}
\subsection{Delayed Feedback Models}
Learning with delayed feedback has received considerable attention in the studies of predicting conversion rate (CVR). Chapelle~\cite{DBLP:conf/kdd/Chapelle14} assumed that the delay distribution is exponential and proposed two generalized linear models for predicting the CVR and the delay time, respectively. However, such a strong hypothesis may be hard to model the delay distribution in practice. To address this issue, ~\cite{DBLP:journals/corr/abs-1802-00255} proposed a non-parametric delayed feedback model for CVR prediction, which exploits the kernel density estimation and combines multiple Gaussian distributions to approximate the actual delay distribution. Moreover, several recent works~\cite{DBLP:journals/corr/abs-2011-11826,DBLP:conf/ijcai/SuZDZYWBXHY20} discretize the delay time by day slot to achieve fine-grain survival analysis for delayed feedback problem. However, one significant drawback of the above methods was that all of them only attempted to optimize the observed conversion information rather than the actual delayed conversion, which cannot fully utilize the sparse positive feedback.

% Learning with delayed feedback has received considerable attention in the studies of predicting conversion rate (CVR). Chapelle~\cite{DBLP:conf/kdd/Chapelle14} assumed that the delay distribution is exponential and proposed two generalized linear models for predicting the CVR and the delay time, respectively. However, such a strong hypothesis may be hard to model the delay distribution in practice. To address this issue, ~\cite{DBLP:journals/corr/abs-1802-00255} proposed a non-parametric delayed feedback model for CVR prediction, which exploits the kernel density estimation and combines multiple Gaussian distributions to approximate the actual delay distribution. Moreover, several recent works~\cite{DBLP:journals/corr/abs-2011-11826,DBLP:conf/ijcai/SuZDZYWBXHY20} discretize the delay time by day slot to achieve fine-grain survival analysis for delayed feedback problem. However, one significant drawback of the above methods was that all of them only attempted to optimize the observed conversion information rather than the actual delayed conversion, which cannot fully utilize the sparse positive feedback. 

\subsection{Unbiased CVR Estimation}
Distinct from previous methods, current mainstream approaches employ the importance sampling method to estimate the real expectation $w.r.t$ another observed distribution~\cite{DBLP:conf/recsys/KtenaTTMDHYS19,DBLP:conf/aaai/YangLHZZZT21,DBLP:conf/kdd/GuSFZZ21,DBLP:conf/www/YasuiMFS20}. Ktena et al.~\cite{DBLP:conf/recsys/KtenaTTMDHYS19} assumes that all samples are initially labeled as negative, then duplicate samples with a positive label and ingest them to the training pipeline upon their conversion. To further model CVR prediction from the biased distribution, they propose two fake negative weighted(FNW) and fake negative calibration(FNC) utilizing importance sampling~\cite{bishop2006pattern}. However, it only focuses on the timeliness of samples and neglects the accuracy of labels. To address this, ES-DFM~\cite{lee2012estimating} introduces an observation window to study the trade-off between waiting for more accurate labels in the window and exploiting fresher training data out of the window. Gu et al.~\cite{DBLP:conf/kdd/GuSFZZ21} further duplicate the real negative and sparse positive in the observation window to eliminate the feature distribution bias introduced by duplicating delayed positive samples. 

% Distinct from previous methods directly optimizing the observed distribution, current mainstream approaches employ the importance sampling method to estimate the real expectation $w.r.t$ another observed distribution~\cite{DBLP:conf/recsys/KtenaTTMDHYS19,DBLP:conf/aaai/YangLHZZZT21,DBLP:conf/kdd/GuSFZZ21,DBLP:conf/www/YasuiMFS20}. Ktena et al.~\cite{DBLP:conf/recsys/KtenaTTMDHYS19} assumes that all samples are initially labeled as negative, then duplicate samples with a positive label and ingest them to the training pipeline upon their conversion. To further model CVR prediction from the biased distribution, they propose two fake negative weighted(FNW) and fake negative calibration(FNC) utilizing importance sampling~\cite{bishop2006pattern}. However, it only focuses on the timeliness of samples and neglects the accuracy of labels. To address this, ES-DFM~\cite{lee2012estimating} introduces an observation window to study the trade-off between waiting for more accurate labels in the window and exploiting fresher training data out of the window. Gu et al.~\cite{DBLP:conf/kdd/GuSFZZ21} further duplicate the real negative and sparse positive in the observation window to eliminate the bias in feature distribution introduced by duplicating delayed positive samples. 

\subsection{Delay Bandits}
The delayed feedback has attracted much attention in bandit methods~\cite{pike2018bandits,mandel2015queue,pike2018bandits}. Previous methods consider the delayed feedback modeling as a sequential decision-making problem and maximum the long-term rewards~\cite{joulani2013online,vernade2017stochastic,heliou2020gradient}. Joulani et al.~\cite{joulani2013online} provided meta-algorithms that transform algorithms developed for the non-delayed, and analyzed the effect of delayed feedback in streaming learning problems. 
% It turns out that delay increases the regret in a multiplicative way in adversarial problems, and in an additive way in stochastic problems. 
\cite{vernade2017stochastic} provided a stochastic delayed bandit model and proof the algorithms in censored and uncensored settings under the hypothesis that the delay distribution is known.
% \cite{grover2018best} introduced a framework in stochastic multi-armed bandit problems with partial and delayed feedback, and particularly designed algorithms for settings where the partial feedback are biased or unbiased estimators of the delayed feedback.
\cite{heliou2020gradient} tried to examine the bandit streaming learning in games with continuous action spaces and introduced a gradient-free learning policy with delayed rewards and bandit feedback. %\cite{DBLP:conf/sigir/ZhangJSWXW21} formulates the streaming recommendation with delayed feedback as a problem of sequential decision making and models it with a batched bandit.

\begin{table}[t]
    \centering
    \caption{Main notations used in the paper.}
    \begin{tabular}{lp{5.5cm}}
        \toprule
        \textbf{Notation} & \textbf{Explanation} \\
        \midrule
        \midrule
        $\textbf{x}$, $y$, $d$ & the input features, label, and interval between click and pay time\\
        $p(\mathbf{x},y), q(\mathbf{x},y)$ & the ground-truth and observed distribution \\
        $w_o$, $w_a$ & the observation and attribution window \\
        $w_i(\mathbf{x},y), z(\mathbf{x})$ & the importance weight and hidden variable of $\mathbf{x}$ \\
        \bottomrule
    \end{tabular}
    \label{tab:notations}
\end{table}

\section{Preliminary}
% ● 基础概念：（附图）
%   ○ odl 归因/观测窗口概念
%   ○ 样本真实分布与观测分布
%   ○ datastream(引用ES-DFM，略讲)及本文的建模目标
% ● 重要性采样背景（详略根据篇幅待定）
In this section, we first formulate the problem of streaming CVR prediction with delayed feedback. Then we give a brief introduction to the standard importance sampling algorithms used in the previous methods. The notations used throughout this paper are summarized in Table~\ref{tab:notations}.

\subsection{Problem Formulation}
% CVR prediction参量，定义
\label{sec:problem_formulation}
In a standard CVR prediction task, the input can be formally defined as $(\mathbf{x},y) \sim p(\mathbf{x},y)$, where $\mathbf{x}$ denotes the features and $y \in \{0, 1\}$ is the conversion label. A generic CVR prediction model aims to learn the parameters $\theta$ of the binary classifier function $f$ by optimizing following ideal loss~\cite{lu2017practical,lee2012estimating}:

\begin{align}
\label{eq:ideal}
\mathcal{L}_{ideal} = \mathbb{E}_{(\mathbf{x},y) \sim p(\mathbf{x},y)} \ell (y,f_{\theta}(\mathbf{x})),
\end{align}
where $(\mathbf{x},y)$ is the training samples drawn from the ground-truth distribution $p(\mathbf{x},y)$, and $\ell $ denotes the classification loss, e.g., the widely used cross-entropy loss. However, as mentioned above, due to the introduction of the observation window, the clicks with conversions that happened outside the observation window will firstly be treated as fake negatives. Thus, the observation distribution $q(\mathbf{x},y)$ is always biased from the ground-truth distribution $p(\mathbf{x},y)$. More specifically, as illustrated in Figure~\ref{fig:delay_click}, there are four types of samples in the online advertising system:

\begin{itemize}[leftmargin=*]
    \item Immediate Positive(IP), e.g.,$d < w_o$. The samples convert inside the observation window are labeled as immediate positive.
    \item Fake Negative(FN), e.g., $w_o < d < w_a$. Fake negative denotes samples that incorrectly labeled as negative at training time due to delay conversion.
    \item Real Negative(RN), e.g., $d > w_a / d = \infty$. The samples not convert after waiting a sufficient long attribution window $w_a$ are labeled as real negative.
    \item Delay Positive(DP). These samples are duplicated and investigated into the training pipeline with a positive label upon conversion.
\end{itemize}

\begin{figure}[tbp]
    \centering
    \includegraphics[width=.9\linewidth]{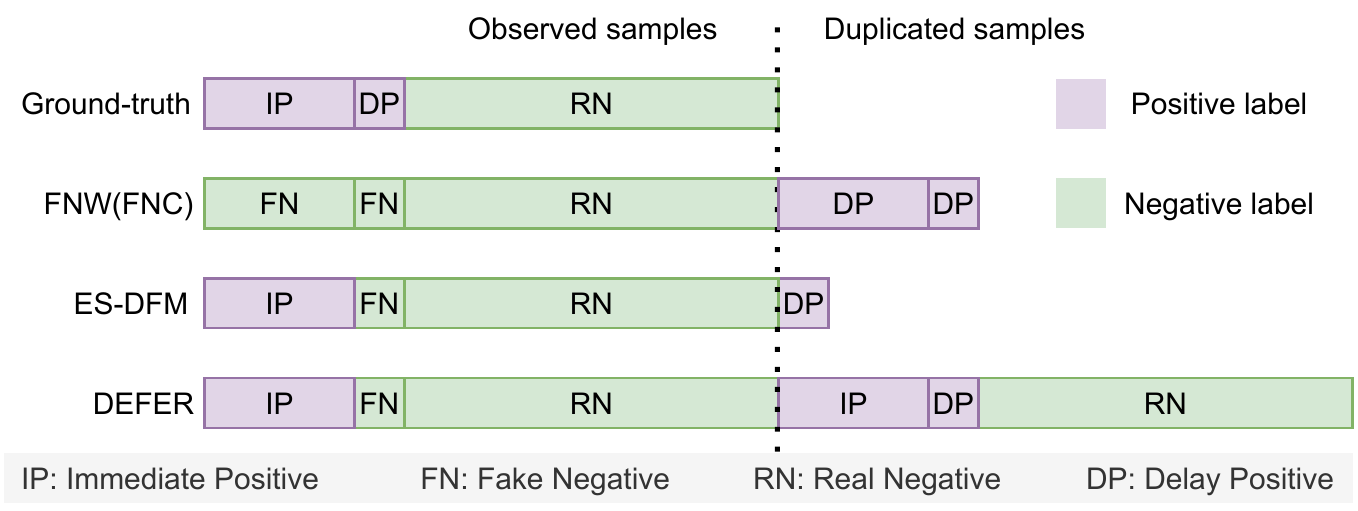}
    \caption{Data Distribution of previous online delayed feedback methods.}
    \label{fig:huibu}
\end{figure}

\subsection{Importance Sampling}

Importance sampling has been widely studied and applied in many recent tasks, e.g., counterfactual learning~\cite{DBLP:conf/icde/XiaoJTLXXY21} and unbiased estimation~\cite{DBLP:conf/aaai/YangLHZZZT21,DBLP:conf/kdd/GuSFZZ21}. Typically, previous approaches use importance sampling to estimate the expectation of training loss from the observed distribution and rewrite the ideal CVR loss function as follows:
\begin{align}
\mathcal{L} &= \mathbb{E}_{(\mathbf{x},y) \sim p(\mathbf{x},y)}  \ell(y, f_{\theta}(\mathbf{x})) \\
&= \mathbb{E}_{(\mathbf{x},y) \sim q(\mathbf{x},y)} w(\mathbf{x},y) \ell(y, f_{\theta}(\mathbf{x})),
\end{align}
where $f_{\theta}$ is the desired CVR model pursuing unbiased CVR prediction, $p(\mathbf{x}, y)$ and $q(\mathbf{x}, y)$ respectively denote the joint density function of the \textit{ground truth} and the \textit{observed} and \textit{duplicated} distribution, and $w(\mathbf{x},y)$ is the likelihood ratio of the \textit{ground truth} distribution with respect to the \textit{observed} and \textit{duplicated} distribution introduced by importance sampling, chasing for an unbiased $f^{\ast}_{\theta}(\mathbf{x})$.

% w(x,y)的计算思路最先引自(Addressing Delayed Feedback for Continuous Training with Neural Networks in CTR prediction)

Currently, by assuming or ensuring $p(\mathbf{x}) \approx q(\mathbf{x})$ and carefully designing the sample duplicating mechanisms, all published methods~\cite{DBLP:conf/recsys/KtenaTTMDHYS19,DBLP:conf/aaai/YangLHZZZT21,DBLP:conf/kdd/GuSFZZ21} apply the same derivation of the formulation of $w(\mathbf{x},y)$ that firstly published at ~\cite{DBLP:conf/recsys/KtenaTTMDHYS19} as follows:
\begin{align}
\mathcal{L} &= \mathbb{E}_{(\mathbf{x},y) \sim q(\mathbf{x},y)}  w(\mathbf{x},y) \ell(y, f_{\theta}(\mathbf{x})) \\
% &= \int\int q(\mathbf{x},y) \ell(y,f_{\theta}(\mathbf{x})) dxdy \\
\label{eq:raw_is}
&= \int \! q(\mathbf{x}) dx \! \int \! q(y\! \mid \! \mathbf{x}) \frac{p(\mathbf{x},y)}{q(\mathbf{x},y)} \ell(\mathbf{x}, y; f_{\theta}(\mathbf{x})) dy \\
&= \int \! q(\mathbf{x}) \frac{p(\mathbf{x)}}{q(\mathbf{x})} dx \! \int \! q(y \! \mid \! \mathbf{x}) \frac{p(y|\mathbf{x})}{q(y|\mathbf{x})} \ell(\mathbf{x}, y; f_{\theta}(\mathbf{x})) dy \\
\label{eq:approx}
&\approx  \int \! q(\mathbf{x}) dx \! \int \! q(y \! \mid \! x) \frac{p(y|\mathbf{x})}{q(y|\mathbf{x})} \ell(\mathbf{x}, y; f_{\theta}(\mathbf{x})) dy \\ 
\label{eq:fix}
&\approx \!\sum_{(\mathbf{x}_i,y_i) \in \mathcal{D}} \Bigl[ y_i \frac{p(y_i\!=\!1\! \mid \!\mathbf{x}_i)}{q(y_i\!=\!1\! \mid \! \mathbf{x}_i)} \log f_{\theta}(\mathbf{x}_i)
\notag\\
&+ (1\!-\!y_i) \frac{p(y_i\!=\!0\! \mid \! \mathbf{x}_i)}{q(y_i\!=\!0 \!\mid \!\mathbf{x}_i)} \log (1\!-\!f_{\theta}(\mathbf{x}_i))\Bigr],
\end{align}
where $\mathcal{D}$ is the observed dataset. The difference among these published methods mainly lies in:
\begin{enumerate}[leftmargin=*]
\item Different designs of the training pipeline, e.g. choice of $w_o$ and the definition of duplicated samples as illustrated in Figure~\ref{fig:huibu}, which eventually results in different formulations of $q(y \! \mid \! \mathbf{x})$
\item Different choices of modelling $p(d>w_o \! \mid \! \mathbf{x},y=1)$ or $p(d>w_o|y=1)p(y=1|x)$, etc.
\end{enumerate}

As demonstrated in Figure~\ref{fig:huibu}, FNW/FNC~\cite{DBLP:conf/recsys/KtenaTTMDHYS19} firstly sets $w_o=0$ and marks all clicks as negative samples at click time and all positive samples are collected and replayed as DP at conversion time; ES-DFM~\cite{DBLP:conf/aaai/YangLHZZZT21} and DEFER~\cite{DBLP:conf/kdd/GuSFZZ21} keep a reasonable observation time $w_o$, hence clicks with conversions happened within $t_0 + w_o$ can be correctly labelled as IP. The only difference between ES-DFM and DEFER is that ES-DFM only replay delayed positives while DEFER duplicates all clicks(including IP and RN). Both ES-DFM and DEFER choose to model $f_{dp}(\textbf{x})=p(d>w_o,y\!=\!1 \! \mid \! \textbf{x})=p(d>w_o \! \mid \! \textbf{x},y\!=\!1)p(y\!=\!1 \! \mid \! \textbf{x})$ as a whole. Such differences of these methods in sample duplicating mechanisms eventually result in their different formulation of $q(y|x)$ as shown in equation~(\ref{eq:fnw}), ~(\ref{eq:esdfm}) and ~(\ref{eq:defer}), respectively.

\begin{align}
    \label{eq:fnw}
    q_{\mathrm{fnw}}(y = 0 \mid \mathbf{x}) &= \frac{1}{1 + p(y=1 \mid \mathbf{x})} \\
    \label{eq:esdfm}
    q_{\mathrm{esdfm}}(y = 0 \mid \mathbf{x}) &= \frac{p(y=0 \mid \mathbf{x}) + f_{dp}(\mathbf{x}) }{1 + f_{dp}(\mathbf{x})} \\
    \label{eq:defer}
    q_{\mathrm{defer}}(y = 0 \mid \mathbf{x}) &= p(y=0 \mid \mathbf{x}) + \frac{1}{2}f_{dp}(\mathbf{x}),
\end{align}

\subsubsection{Limitations}
\label{sec:limitation}
Despite their success in bias reduction, we notice that these published methods still failed to achieve unbiased CVR prediction due to a hidden flaw introduced while deriving the formulation of $w(\mathbf{x},y)$.
Normally, importance sampling assumes no value modification during the transition from $p(\mathbf{x},y)$ to $q(\mathbf{x},y)$, whereas in CVR prediction as mentioned in Section~\ref{sec:problem_formulation}, even for the same clicks, observed label from $q(\mathbf{x},y)$ can be temporarily deviated to ground-truth label from $p(\mathbf{x},y)$. 
More specifically and rigorously, if we distinguish the observed label as $v$ and re-express the biased distribution as $q(\mathbf{x},v)$, we have:
\begin{align}
\label{eq:bias_sample}
y = y(v,d)=\begin{cases}
 1, \quad v=1\\
 0, \quad v=0, d=+\infty \\
 1, \quad v=0, d > w_o.
\end{cases}
\end{align}

As a result, fake negative samples which should be denoted as $\frac{p(d > w_o, y=1 \mid \mathbf{x})}{q(y=0 \mid \mathbf{x})}$, are falsely treated as real negative in equation~(\ref{eq:fix}), leading to suboptimal performance and biased CVR prediction.

\begin{table}[t]
    \centering
    \caption{The label disparsity of observed distribution and actual distribution after attribution in terms of four sample types. The observability denotes whether the model can distinguish the sample type during streaming training.}
    \begin{tabular}{c|cccc}
        \toprule
         & IP & FN & RN & DP  \\
         \midrule
         \midrule
        observability & True & False & False & True \\
        observed label & 1 & 0 & 0 & 1 \\
        attributed label & 1 & 1 & 0 & 1 \\
        \bottomrule
    \end{tabular}
    \label{tab:label_diff}
\end{table}
\section{Methodology}
% p1: delay建模总体介绍
In this section, we present our proposed method DElayed Feedback modeling with UnbiaSed Estimation~(DEFUSE) in detail. We First introduce our correction of unbiased estimation, which respectively weight the importance of four types of samples. We then propose a two-step optimization for DEFUSE. Finally, to further reduce the influence caused by the observed bias distribution, we devise a bi-distribution modeling framework to sufficiently utilize the immediate conversion under the actual distribution. Note that our DEFUSE is applicable for different training pipelines, but for ease of description, we will introduce our approach on top of the design of the training pipeline in ES-DFM.

\subsection{Unbiased Delayed Feedback Modeling}
% 重要性采样修正
As we describe in Section~\ref{sec:limitation}, our goal is to achieve unbiased delayed feedback modeling by further optimizing the loss of the fake negative samples. From Equation~(\ref{eq:raw_is},\ref{eq:bias_sample}), we can obtain the unbiased estimation as:
\begin{align}
\label{eq:ub}
\mathcal{L}_{ub} &= \! \int \! q(\mathbf{x}) dx \! \int \! q(v \! \mid \! x) \frac{p(\mathbf{x})}{q(\mathbf{x})}
\frac{p(y(v,d)|\mathbf{x})}{q(v|\mathbf{x})} \ell(\mathbf{x}, y(v,d); f_{\theta}(\mathbf{x})) dv.
\end{align}
% \begin{align}
% % \mathcal{L}_{ideal} &= \mathbb{E}_{(\mathbf{x},y) \sim p(\mathbf{x},y)}  \ell(y, f_{\theta}(\mathbf{x}) \\
% % &= \int\int p(\mathbf{x},y) \ell(y,f_{\theta}(\mathbf{x})) dxdy \\
% \label{eq:ub}
% \mathcal{L}_{ub} &= \! \int \! q(\mathbf{x}) dx \! \int \! q(v \! \mid \! x) \frac{p(\mathbf{x})}{q(\mathbf{x})}
% \frac{p(y(v,d)|\mathbf{x})}{q(v|\mathbf{x})} \ell(\mathbf{x}, y(v,d); f_{\theta}(\mathbf{x})) dv.
% % \\
% % \label{eq:lub}
% % &=\sum_{\mathbf{x}_i,y_i} \frac{p(\mathbf{x}_i)}{q(\mathbf{x}_i)} (1+f_{dp}(\mathbf{x})) \left[  y_i \ell_{pos} + (1-y_i) \ell_{neg} \right] \\
% % &=\mathcal{L}_{ub},
% \end{align}

Where $\ell(\mathbf{x}, y(v,d); f_{\theta}(\mathbf{x}))$ is the loss function for observed samples with label $y(v,d)$. 
Typically, Previous approaches eliminate $\frac{p(\mathbf{x})}{q(\mathbf{x})}$ by assuming $p(\mathbf{x}) \approx q(\mathbf{x})$~\cite{DBLP:conf/recsys/KtenaTTMDHYS19,DBLP:conf/aaai/YangLHZZZT21} or design proper training pipeline to guarantee equal feature distribution~\cite{DBLP:conf/kdd/GuSFZZ21}. 

% $\ldots$, we directly model $\frac{p(\mathbf{x})}{q(\mathbf{x})}$ as:
% %在联合分布下的比重
% \begin{align}
% \frac{p(\mathbf{x})}{q(\mathbf{x})} &= \frac{p(\mathbf{x})}{p(\mathbf{x}) + p(\mathbf{x}, y=1, d > w_1)} = \frac{1}{1 + f_{dp}(\mathbf{x})},
% \end{align}
% where $f_{dp}(\mathbf{x})$ is the delay positive probability, denotes the probability that a sample is a duplicated positive.

\subsubsection{Importance weighting of DEFUSE}

In this work, different from previous works that focus on the duplicating mechanism, we aim for unbiased CVR estimation by properly evaluating the importance weight for $\ell(\mathbf{x}, y(v,d); f_{\theta}(\mathbf{x}))$. 
As shown in Table~\ref{tab:label_diff}, the observed samples can be formally divided into four parts. Intuitively, if we have all labels of each part, the Equation~(\ref{eq:ub}) can be rewritten as:
\begin{align}
\label{eq:ub2}
\mathcal{L}_{ub} \! &= \! \int \! q(\mathbf{x}) \Bigl[\sum_{v_i} q(v_i \! \mid \! x) w_i(\mathbf{x}, y(v_i, d)) \ell(\mathbf{x}, y(v_i,d); f_{\theta}(\mathbf{x})) \Bigr] dx,
\end{align}
where $w_i=\frac{p(\mathbf{x}, y(v_i,d))}{q(\mathbf{x}, v_i)}$ and $i \in \{IP, FN, RN, DP \}$, subject to $\sum {v_i}=1$ and $v_i \in \{0,1\}$.
Note that present works merely model the observed positives and negatives in Equation~(\ref{eq:fix}), which ignore the impact of fake negatives(FN) and lead to bias in label distribution. To solve equation~(\ref{eq:ub2}), we first introduce a latent variable $z$, which is used to inference whether an observed negative is FN or not, then respectively modeling the importance weights $w_i$ of these four types of observed samples. Thus, equation~(\ref{eq:ub2}) is equivalent to: 
\begin{align}
\label{eq:new_ub}
& \min_{\theta} \mathcal{L}_{ub} \notag \\
\Leftrightarrow & \min_{\theta} \! \int \! q(\mathbf{x}) \Bigl[v(w_{DP}log f_\theta(\mathbf{x}) + \mathbb{I}_{IP}(w_{IP}-w_{DP})\log f_\theta(\mathbf{x})) \notag\\
&+ (1-v)(w_{FN}\log f_\theta(\mathbf{x})z + w_{RN}\log (1-f_\theta(\mathbf{x}))(1-z)) \Bigr] dx
\end{align}
%In this work, different from previous works that focus on the duplicating mechanism, we are aiming for unbiased CVR estimation by properly evaluate the importance weight for $\ell(\mathbf{x}, y(v,d); f_{\theta}(\mathbf{x}))$. 
%As shown in Table~\ref{tab:label_diff}, the observed samples can be formally divided into four parts. 
%However, present works merely model the observed positive and negative in Equation~(\ref{eq:fix}), which ignores the impact of fake negatives and leads to bias in label distribution. To further reduce the bias, we respectively modeling the importance weights of these four types of observed samples. Formally, we reformulate the 
% $\mathcal{F}_{iw} = \frac{p(y(v,d)|\mathbf{x})}{q(v|\mathbf{x})} \ell(\mathbf{x}, v; f_{\theta}(\mathbf{x}))$in 
%Equation~(\ref{eq:ub}) as:

% 还需要减去loss5？

% \begin{align}
% \label{eq:new_ub}
% \mathcal{L}_{ub} =& \sum y v w_{IP} \log (f_{\theta}(\mathbf{x})p(d \leq w_o)) \notag \\
% &+ \sum \left[ y(1-v)w_{DP} + (1-y)z_i w_{FN} \right] \log (f_{\theta}(\mathbf{x}) p(d>w_o)) \notag \\
% &+ \sum (1-y)(1-z_i) w_{RN} \log (1-f_{\theta}(\mathbf{x}))
% \end{align}
% \begin{align}
%     \label{eq:new_ub}
%     \mathcal{L}_{ub} &= - \sum_{x,v} v(w_P(\mathbf{x}) + w_{DP}(\mathbf{x})) \log \left(f_{\theta}(\mathbf{x})\right) 
%     \notag 
%     \\&- \sum_{x,v} (1-v)(w_{FN}(\mathbf{x}) + w_{RN}(\mathbf{x})) \log \left(1 - f_{\theta}(\mathbf{x}) + f_{\theta}(\mathbf{x})f_{dp}^{\prime}(\mathbf{x})\right) 
% \end{align}
$s.t.$ 
\begin{align*}
    w_{IP}(\mathbf{x}) = w_{RN}(\mathbf{x}) &= 1 + f_{dp}(\mathbf{x}) \\
    w_{DP}(\mathbf{x}) + w_{FN}(\mathbf{x}) &= 1 + f_{dp}(\mathbf{x}),
\end{align*}
where $w_{IP}(\mathbf{x})$, $w_{DP}(\mathbf{x})$, $w_{FN}(\mathbf{x})$, $w_{RN}(\mathbf{x})$ denotes the importance weights; $\mathbb{I}_{IP}$ is the indicator of observed immediate positives. Empirically, we set $w_{DP}(\mathbf{x})\!=\!1$ and $w_{FN}(\mathbf{x})\!=\!f_{dp}(\mathbf{x})$ since $DP$ can be observed. A detailed proof is given in the supplementary material. 
%where $w_{IP}(\mathbf{x})$, $w_{DP}(\mathbf{x})$, $w_{FN}(\mathbf{x})$, $w_{RN}(\mathbf{x})$ denotes the importance weights; $y$, $v$ respectively denotes the ground-truth and observed label. $z_i$ is used to indicate whether an observed negative to be FN. A detailed proof is given in the supplementary material. 

% where $\ell_{pos} = \log f_{\theta}$ and $\ell_{neg} = \log (1-f_{\theta} + f_{dp}(\mathbf{x}))$. Similarly, by using Equation~(\ref{eq:esdfm}), we have:

% Then according to Equation~(\ref{eq:lub}), we can obtain the final unbiased estimation of importance sampling as:
% \begin{align}
%     \label{eq:ub}
%     \mathcal{L}_{ub}=\sum_{\mathbf{x}_i,y_i} \left[ y_i \log f_{\theta}(\mathbf{x}) + (1-y_i) \log (1 - f_{\theta}(\mathbf{x}) + f_{dp}(\mathbf{x}) \right].
% \end{align}
Compared with the standard cross-entropy loss, we integrate an auxiliary task $f_{dp}(\mathbf{x})$ to model the importance weights $w_i$ for each type of sample, rather than directly using the observed label. 
\subsubsection{Optimization}
\label{sec:opt}
Hereafter, the remaining question is how to optimize the unbiased loss function. 
% More specifically, we employs 
% The remain question Although theoretically unbiased, the drawback of this joint optimization algorithm is that the multiplicative term $f_{\theta}(\mathbf{x}) f_{dp}^{\prime}(\mathbf{x})$ may cause high variance. This typically implies slow convergence and leads to suboptimal performance. 
Since the $z$ is inaccessible in Equation~(\ref{eq:new_ub}), we implement a two-step optimization by introducing another auxiliary model $z(\mathbf{x})$ predicting the hidden $z$ to further decouple the observed negative samples into real negative and fake negative samples:
\begin{align}
    % \mathcal{L}_{neg} &= w_i \frac{f_{dp}(\mathbf{x})}{p(y=0) +f_{dp}(\mathbf{x})} \log f_{dp}^{\prime}(\mathbf{x}) f_{\theta}(\mathbf{x}) \\
    % &+ (1-w_i) \frac{p(y=0)}{p(y=0) + f_{dp}(\mathbf{x})} \log(1 - f_{\theta}(\mathbf{x})
    \label{eq:em}
    \mathcal{L}_{neg} =& z(\mathbf{x}) w_{FN} \log f_{\theta}(\mathbf{x})
    + (1 - z(\mathbf{x})) w_{RN} \log (1 - f_{\theta}(\mathbf{x}))
\end{align}
where
\begin{align}
    \label{eq:z_prob}
    z(\mathbf{x}) = \frac{p(y=1, d > w_o \mid \mathbf{x})}{p(y=0 \mid \mathbf{x}) + p(y=1, d > w_o \mid \mathbf{x})},
\end{align}
where $z(\mathbf{x})$ is the fake negative probability, denotes the probability that an observed negative is a ground truth positive.

\begin{figure}[tbp]
    \centering
    \begin{subfigure}[b]{0.68\linewidth}
        \centering
        \includegraphics[width=\textwidth]{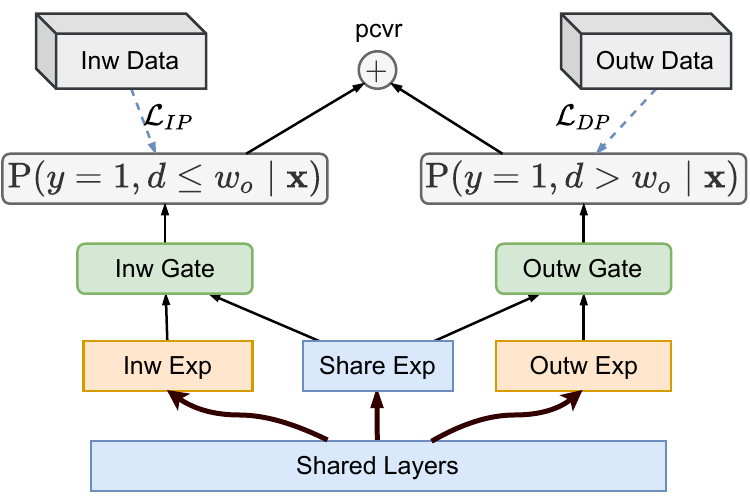}
        \caption{Model architecture.}
        \label{fig:bi-dis}
    \end{subfigure}
    \begin{subfigure}[b]{0.28\linewidth}
        \centering
        \includegraphics[width=\textwidth]{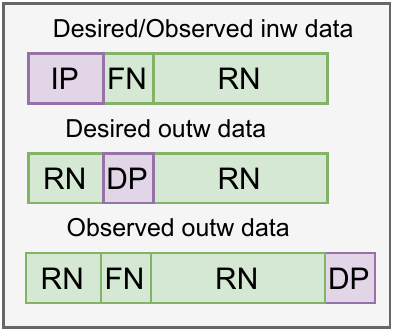}
        \caption{Labeled data.}
        \label{fig:bi-dis2}
     \end{subfigure}
    \caption{An illustration of Bi-Distribution model, where Exp, Share Exp, Outw Exp respectively denotes the single layer in\_window, shared, and out\_window expert network.}
\end{figure}

In practice, we implement two ways to model the $z(\mathbf{x})$:
\begin{itemize}[leftmargin=*]
    \item $z_{1}(\mathbf{x}) = 1 - f_{rn}(\mathbf{x})$. This adopts a binary classification model $f_{rn}(\mathbf{x})$ to predict the probability of an observed negative being a real negative~\cite{DBLP:conf/aaai/YangLHZZZT21}. For the training of $f_{rn}$ model, the observed positives are excluded, then the negatives are labeled as 1 and the delayed positives are labeled as 0. % 直接二分类
    \item $z_{2}(\mathbf{x}) = \frac{f_{dp}(\mathbf{x})}{f_{dp}(\mathbf{x}) + 1 - f_{\theta}(\mathbf{x})}$. This adopts the CVR model $f_{\theta}(\mathbf{x})$ and delay model $f_{dp}(\mathbf{x})$ to indirectly model the fake negative probability. For the learning of $f_{dp}(\mathbf{x})$, the delayed positives are labeled as 1, the others are labeled as 0.
    % \item $w_{i,3}(\mathbf{x}) = \frac{1 - f_{\theta}(\mathbf{x}).  }{f_{\theta}(\mathbf{x})f_{dp}^{\prime}(\mathbf{x}) + 1 - f_{\theta}(\mathbf{x})}$. Since the $f_{rn}(\mathbf{x})$ and $f_{dp}(\mathbf{x})$ is hard to train during continuous learning settings due to the delayed feedback problem, $w_{i,3}(\mathbf{x})$ utilize the $f_{dp}^{\prime}(\mathbf{x})$ to ensure the freshness.
\end{itemize}
% Finally, considering Equation~(\ref{eq:em}), the importance sampling loss is:
% \begin{align}
%     \mathcal{L}_{ub,EM} &= \sum_{\mathbf{x}_i,y_i} y_i \log f_{\theta}(\mathbf{x}) + (1-y_i) \mathcal{L}_{neg,EM}.
% \end{align}

\subsection{Bi-Distribution Modeling}
% 想个好听点的名字？
Although theoretically unbiased, a potential drawback of our DEFUSE is that the estimation of importance weights $w$, hidden model $z(\mathbf{x})$ and especially the multiplicative term $z(\mathbf{x}) w_{FN}$ and $(1-z(\mathbf{x}))w_{RN}$ may cause high variance. 
This typically implies slow convergence and leads to sub-optimal performance, especially when the feedback is relatively sparse.
As such, we strive to build an alternative learning framework that can fully exploit samples from the observed distribution directly.
 
Recall that, distinct from previous methods merely using the observed positive and negative samples, we divide the samples into four types. The IP and DP denote the immediate conversion and the delay conversion, respectively.

We thus adopt an multi-task learning~~\cite{DBLP:conf/kdd/MaZYCHC18,DBLP:conf/icde/XiaoJTLXXY21,DBLP:conf/kdd/PanMRSF19,DBLP:conf/kdd/LeeODL12} framework to jointly optimizing following subtasks: 1) In window(Inw) model: predicting the IP probability $\mathcal{F}_{IP}(\mathbf{x}) = p(y=1, d \leq w_o \mid \mathbf{x})$ within the observation window $w_o$. 2) Out window(Outw) model: predicting the DP probability $\mathcal{F}_{DP}(\mathbf{x})$ out of $w_o$. Then the overall conversion probability can be formalized as:
\begin{align}
    p(y=1 \mid \mathbf{x}) = \mathcal{F}_{IP}(\mathbf{x}) + \mathcal{F}_{DP}(\mathbf{x}).
\end{align}

It's worth mentioning that, as demonstrated in Figure~\ref{fig:bi-dis2}, for task 1), the samples used are nonduplicate and correctly labeled, so the $\mathcal{F}_{IP}(\mathbf{x})$ model can be directly trained upon the ground-truth distribution without bias. For task 2), the $\mathcal{F}_{DP}(\mathbf{x})$ model has to be trained on a biased observed distribution identical to that of FNW~\cite{DBLP:conf/recsys/KtenaTTMDHYS19} with $w'_{o}=0$. 
%For the prediction of $\mathcal{F}_{DP}(\mathbf{x})$, since the coexistence of FN and DP results in the same samples marked with different labels, 
Thus, we implement our DEFUSE to the $\mathcal{F}_{DP}(\mathbf{x})$ model to achieve unbiased estimation with importance sampling. Similar to the derivation of equation~(\ref{eq:new_ub}), we have:
\begin{align}
    %\mathcal{L}_{IP} &= \sum_{(x,y_{IP}) \in \mathcal{D}_{1}} y_{IP} \log f_{IP}(\mathbf{x}) + (1 - y_{IP}) \log (1 - f_{IP}(\mathbf{x})) \\
    \mathcal{L}_{IP} &= \int p(\mathbf{x},y_{IP}) \Bigl[ y_{IP} \log f_{IP}(\mathbf{x}) + (1 - y_{IP}) \log (1 - f_{IP}(\mathbf{x}))\Bigr] dx \\
    \mathcal{L}_{DP} &= \int q(\mathbf{x},v_{DP}) \Bigl[v_{DP} w^{\prime}_{DP}(\mathbf{x}) \log f_{DP} (\mathbf{x}) \notag\\
    &+ 
    % \sum_{(x,y_{DP}) \in \mathcal{D}_{2}}
    (1 - v_{DP})w^{\prime}_{FN}(\mathbf{x}) z^{\prime}(\mathbf{x}) \log f_{DP}(\mathbf{x}) \notag\\
    &+ 
    % \sum_{(x,y_{DP}) \in \mathcal{D}_{2}}
    (1 - v_{DP}) w^{\prime}_{RN}(\mathbf{x}) (1 - z^{\prime}(\mathbf{x})) \log (1 - f_{DP}(\mathbf{x}))
    \Bigr]dx,
    % \mathcal{L}_{DP} &= \sum_{(x,y_{DP}) \in \mathcal{D}_{2}}y_{DP} w^{\prime}_{DP}(\mathbf{x}) \log f_{DP} (\mathbf{x}) \notag\\
    % &+ 
    % % \sum_{(x,y_{DP}) \in \mathcal{D}_{2}}
    % (1 - y_{DP})w^{\prime}_{FN} z_i^{\prime}(\mathbf{x}) \log f_{DP}(\mathbf{x}) \notag\\
    % &+ 
    % % \sum_{(x,y_{DP}) \in \mathcal{D}_{2}}
    % (1 - y_{DP}) w^{\prime}_{RN} (1 - z_i^{\prime}(\mathbf{x})) \log (1 - f_{DP}(\mathbf{x})),
\end{align}
$s.t.$ 
\begin{align}
    w^{\prime}_{DP}(\mathbf{x}) + w^{\prime}_{FN}(\mathbf{x}) = 1 + f_{dp}(\mathbf{x}), \quad    w^{\prime}_{RN}(\mathbf{x}) = 1 + f_{dp}(\mathbf{x}), \notag 
\end{align}
\begin{table*}[t]
    \centering
    \caption{Statistic of Criteo and Taobao Dataset.}
    \begin{tabular}{c|c|c|c|c|c|c|c}
        \toprule
         Dataset & \#Users & \#Items & \#Features & \#Conversions & \#Samples & avg CVR & Duration\\
         \midrule
         Criteo & - & 5443 & 17 & 3619801 & 15898883 & 0.2269 & 60 days\\
         Taobao Dataset & 382 million & 10.6 million & 23 & 208 million & 5.2 billion & 0.04005 & 21 days \\
         \bottomrule
    \end{tabular}
    \label{tab:statistic}
\end{table*}
where $p(x, y_{IP})$, $q(\mathbf{x}, v_{DP})$ respectively denote distributions of the training datasets for the sub-tasks, $w^{\prime}_{DP}(\mathbf{x})$, $w^{\prime}_{FN}(\mathbf{x})$, and $w^{\prime}_{RN}(\mathbf{x})$ are the importance weights, and $z^{\prime}(\mathbf{x})$ as the hidden model to further infer fake negatives. Finally, we devise our multi-task learning architecture as illustrated in Figure~\ref{fig:bi-dis} to learn the desired CVR model by jointly optimizing the union loss:
\begin{align}
    \mathcal{L} = \mathcal{L}_{IP} + \mathcal{L}_{DP}.
\end{align}

By doing so, we divide the delayed feedback modeling into an unbiased $in\_window$ prediction and an importance sampling-based $out\_window$ prediction task. Note that only the second part needs to be trained with importance weights and hidden variable $z$, which implies that the negative impact introduced by the high variance of inferring $w$ and $z$ can be effectively limited.

% \begin{figure}[tbp]
%     \centering
%     \includegraphics[width=.7\linewidth]{Figures/huibu_fnw_vs_outw.pdf}
%     \caption{The distribution of in/out window dataset.}
%     \label{fig:bi-dis}
% \end{figure}

\section{Experiments}
In this section, we first describe experimental settings and then conduct experiments on both public and industry advertising datasets to evaluate our proposed model by answering the following research questions:
\begin{itemize}[leftmargin=*]
    \item \textbf{RQ1} How does DEFUSE perform on streaming CVR prediction tasks as compared to other state-of-the-art methods?
    \item \textbf{RQ2} How does DEFUSE perform under different duplicating mechanisms?
    \item \textbf{RQ3} How do different components (e.g., hidden variable estimation, observation window size) and hyper-parameter settings affect the results of DEFUSE?
\end{itemize}

\subsection{Datasets}
We evaluate our experiment on both public and industrial datasets. The statistics of the processed datasets are shown in Table~\ref{tab:statistic}.
\begin{itemize}[leftmargin=*]
    %todo
    \item \textbf{Criteo}~\footnote{\url{https://labs.criteo.com/2013/12/conversion-logs-dataset/}} is the well-researched public dataset for the delayed feedback modeling task~\cite{DBLP:conf/kdd/Chapelle14, DBLP:conf/recsys/KtenaTTMDHYS19,DBLP:conf/kdd/GuSFZZ21}. It is collected from the Criteo live traffic data in a period of 60 days with 30 days attribution window. we use the click and pay(if exists) timestamps and all the hashed categorical features and continuous features for train and evaluation. In particular, since 30 days attribution period is unbearable for industrial online advertising, we further derive a one-day attribution version, namely Criteo-1d, which uses samples that convert within one day as positive.
    \item \textbf{Taobao Dataset} is collected from the daily click and conversion logs in Taobao systems. The industrial dataset contains about 5.2 billion interactions between nearly 400 million users and 10 million items. We set $w_a = 1$ day to wait for the actual label for each sample.
    
\end{itemize}

\subsubsection{Data Stream}
% todo
We divide each dataset into two parts to simulate the streaming training environment. 
Specifically, the first shuffled part is used for pre-training a well-initiallized model. 
To prevent label leakage, we refer to the practice of \cite{DBLP:conf/kdd/GuSFZZ21} and set the labels as 0 if the conversion occurs in the second part of the data. For the second part, observed samples are sorted by click time except that the delayed and duplicated samples are ordered by conversion time. We then divide the data into pieces by hour. To simulate the online streaming, we train models on the $t$-th hour data and test them on the $t + 1$-th hour. The reported metrics are the weighted average across different hours on streaming data.

\subsection{Experimental Setting}
\subsubsection{Evaluation Metrics}
% AUC PR-AUC NLL
We applied three widely used evaluation metrics to evaluate the streaming CVR prediction performance: 
\begin{itemize}[leftmargin=*]
    \item \textbf{AUC} is the area under ROC curve which assesses the pairwise ranking performance of the classification results between the conversion and non-conversion samples. 
    \item \textbf{PR-AUC} is the area under the precision-recall curve, which is more sensitive than AUC in skewed data for CVR prediction task.
    \item \textbf{NLL} is originally used in DFM~\cite{DBLP:conf/kdd/Chapelle14}, which is sensitive to the absolute value of the CVR prediction. In a CPA model, the predicted probabilities are important since they are directly used to compute the value of an impression.
\end{itemize}
To demonstrate the relative improvement over the pretrained model, follow previous works~\cite{DBLP:conf/aaai/YangLHZZZT21,DBLP:conf/kdd/GuSFZZ21}, we also evaluate the RI-AUC as:
\begin{align*}
    \mathrm{RI\!-\!AUC_{DEFUSE} = \frac{AUC_{DEFUSE}-AUC_{Pre\!-\!trained}}{AUC_{Oracle}-AUC_{Pre\!-\!trained}}} \times 100\%.
\end{align*}
This indicates the relative improvements for DEFUSE. Obviously, the closer the relative improvement to 100\%, the better the method performs.
\begin{table*}[t]
    \centering
    \caption{The overall performance comparison in terms of AUC, RI-AUC, PR-AUC, and NLL. The results in terms of RI-PR-AUC and RI-NLL are omitted due to space constraints and similar trends. Bold indicates the top-performing method.}
    \begin{tabular}{l|r|r|r|r|r|r|r|r|r|r|r|r}
         \toprule
        % \multirow{2}{*}{Methods}
         Dataset & \multicolumn{4}{c|}{Criteo-30d} & \multicolumn{4}{c|}{Criteo-1d} & \multicolumn{4}{c}{Taobao Dataset} \\
         % \cline{3-8}
         \midrule
         Metrics & AUC & RI-AUC & PR-AUC & NLL & AUC & RI-AUC & PR-AUC & NLL & AUC & RI-AUC & PR-AUC & NLL  \\
         \midrule
         \midrule
         Pre-trained & 0.8307 & 0\% & 0.6251 & 0.4009 & 0.8285 & 0\% & 0.5218 & 0.3001 & 0.8015 & 0\% & 0.6074 & 0.1516 \\
         Vanilla & 0.8098 & -108.29\% & 0.5902 & 0.5453 & 0.8384 & 53.38\% & 0.5408 & 0.3088 & 0.8047 & 32.65\% & 0.6168 & 0.1504 \\
         Vanilla-Win & 0.8375 & 35.23\% & 0.6288 & 0.4055 & 0.8385 & 52.91\% & 0.5408 & 0.3088 & 0.8059 & 44.90\% & 0.6169 & 0.1495 \\
         Oracle & 0.8500 & 100\% & 0.6468 & 0.3869 & 0.8474 & 100\% & 0.5519 & 0.2874 & 0.8113 & 100\% & 0.6197 & 0.1470 \\
        %  \midrule
        %  Pre-trained & \\
        %  Vanilla &  \\
         \midrule
        %  DFM & 0.807 \\
         FNC & 0.8373 & 34.20\% & 0.6222 & 0.4688 & 0.8343 & 30.69\% & 0.4806 & 0.3145 & 0.8053 & 38.78\% & 0.6150 & 0.1495 \\
         FNW & 0.8376 & 35.75\% & 0.6310 & 0.3971 & 0.8348 & 33.33\% & 0.4982 & 0.3367 & 0.8054 & 39.80\% & 0.6148 & 0.1497  \\
         ES-DFM & \underline{0.8396} & \underline{46.11\%} & \underline{0.6384} & \underline{0.3947} & 0.8459 & 92.06\% & \underline{0.5492} & \textbf{0.2885} & 0.8066 & 52.04\% & 0.6155 & \underline{0.1494} \\
         DEFER & 0.8382 & 38.86\% & 0.6338 & 0.4800 &0.8463 & 94.17\% & 0.5490 & 0.3098 & 0.8065 & 51.02\% & 0.6153 &0.1529 \\
         \midrule
         DEFUSE & \textbf{0.8408} & \textbf{52.33\%} & \textbf{0.6400} & \textbf{0.3946} & \underline{0.8465} & \underline{95.24\%} & 0.5490 & \underline{0.3086} & \underline{0.8069} & \underline{55.10\%} & \underline{0.6177} & \textbf{0.1489} \\
         Bi-DEFUSE & 0.8379 & 37.31\% & 0.6301 & 0.3963 & \textbf{0.8467} & \textbf{96.30\%} & \textbf{0.5499} & 0.3092 & \textbf{0.8080} & \textbf{66.33\%} & \textbf{0.6185} & 0.1512 \\
        %  \%Improv.  \\
         \bottomrule
    \end{tabular}
    \label{tab:perform}
\end{table*}

\subsubsection{Baselines} We compared our DEFUSE with the following state-of-the-art methods:
\begin{itemize}[leftmargin=*]
    \item \textbf{Pre-trained}: This model is trained by the first part of data but without continuous training on the streaming data. The rest methods are all finetuned on top of this model during streaming simulation.
    \item \textbf{Oracle}: A model finetuned with the ground-truth label other than the observed label, which denotes the upper bound of the delayed feedback modeling.
    \item \textbf{Vanilla}: A model training with the waiting window but without any duplicate samples, using standard cross-entropy loss.
    \item \textbf{Vanilla-Win}: Vanilla-Win is trained on the streaming data with a waiting window. DP samples are duplicated with the actual label and re-sent to the training pipeline after conversion.
    % \item \textbf{DFM}~\cite{DBLP:conf/kdd/Chapelle14}: A model finetuned on top of the pre-trained model using delayed feedback loss.
    \item \textbf{FNW}~\cite{DBLP:conf/recsys/KtenaTTMDHYS19}: A model finetuned on top of the Pre-trained model using the fake negative weighted loss.
    \item \textbf{FNC}~\cite{DBLP:conf/recsys/KtenaTTMDHYS19}: A model finetuned on top of the Pre-trained model using the fake negative calibration loss.
    \item \textbf{ES-DFM}~\cite{DBLP:conf/aaai/YangLHZZZT21}: It is trained on the same streaming data as Vanilla-Win but introduces auxiliary tasks and using the ES-DFM loss.
    \item \textbf{DEFER}~\cite{DBLP:conf/kdd/GuSFZZ21}: This model is trained on the DEFER pipeline as illustrated in Figure~\ref{fig:huibu} with the DEFER loss.
\end{itemize}
We also tried \textbf{DFM}~\cite{DBLP:conf/kdd/Chapelle14} but found that the delayed feedback loss is difficult to converge on our sizeable industrial dataset due to the difficultly of estimating the delay time based on a strong distribution assumption. Hence, although it achieved promising performance in Criteo, we did not select it for comparison.
\subsubsection{Parameter Settings}
We implement the DEFUSE in Tensorflow. For a fair comparison, we tune the parameter settings of each model. The hidden units are fixed for all models with hidden size $\{256, 256, 128\}$. The Leaky ReLU~\cite{maas2013rectifier} and BatchNorm layer~\cite{DBLP:conf/icml/IoffeS15} are attached to each hidden layer. All methods are trained with Adam~\cite{DBLP:journals/corr/KingmaB14} for optimization.
For the basic parameter settings of all models, we apply the grid search strategy to tune the coefficient of $L_2$ normalization in $\{0.0001, 0.0005, 0.001, 0.01 \}$ and search the learning rate among $\{0.0001, 0.0005, 0.001\}$, or directly copy the best parameter settings reported in the original papers~\cite{DBLP:conf/recsys/KtenaTTMDHYS19,DBLP:conf/aaai/YangLHZZZT21}. Moreover, we tune the waiting window in $\{0.25, 0.5, 1\}$ hour. The same pretrained model is used to initialize the online models.

\subsection{Performance Comparison (RQ1)}

% 结论：
% 1. DEFUSE效果显著
% 2. 横向比较Criteo-30d和Criteo-1d：
% 2.1. Criteo-30d, Vanilla < Pretrain; Criteo-1d, Vanilla > Pretrain，Oracle 在30d归因上更显著: 长归因周期引入更多的FN, 直接在观测分布上使用交叉熵，周期越短收益越高，但是考虑到label准确性，对于归因周期也需要balance。
% 3. 纵向比较：
% 3.1. 在多数case上，FNC/FNW<ES-DFM,验证了观测窗口可以较好的平衡label准确性和时效性。
% 3.2. DEFER不符合预期，这里抛出loss的问题和我们的修正。(问题，Criteo上DEFER修正后仍然不佳，CVR上效果还可以）
% 3.3. DEFUSE1/2/3相对于FNC,ES-DFM,DEFER分别拿到收益。说明本文提出的与分布无关的无偏估计的效果显著。
% 3.4. Bi-DEFUSE在CVR和单天Criteo上效果好而在30d效果差，这是因为对于较短的归因周期，窗口内建模所能利用的真实分布样本更丰富，窗口外建模因高方差带来的影响更小。
To demonstrate the overall performance of DEFUSE, we conduct 5 random runs on the Criteo and Taobao Dataset, and report the average results of all methods in Table~\ref{tab:perform}. The best-performing method is boldfaced. Analyzing such performance comparison, we have the following observations:

\begin{itemize}[leftmargin=*]
    \item Our approach consistently yields significant improvements on all the datasets. In particular, DEFUSE and Bi-DEFUSE improves over the strongest baselines $w.r.t.$ RI-AUC by 6.22\%, 2.13\%, and 15.31\% in Criteo-30d, Criteo-1d, and Taobao Dataset, respectively. Unlike previous approaches that only utilize the observed positive and negative samples during importance sampling, we divide the observed distribution into four types of feedbacks and introduce an auxiliary task to infer fake negatives from the observed negatives. Moreover, on Criteo-1d, our method can narrow the delayed feedback gap significantly compared to other methods by comparing the relative metrics, RI-AUC. Note that as reported in ~\cite{zhou2018deep}, a small improvement of offline AUC can lead to a significant increase in streaming CTR. In our scenario, even 0.1\% of AUC improvement in CVR prediction is substantial and achieves significant online promotion.
    \item ES-DFM and DEFER generally achieve better performance than FNW and FNC. Such improvement can be attributed to the duplicating mechanism with a properly tuned $w_o$, which provides a good balance for the trade-off between label accuracy and sample freshness. Comparison between Vanilla-Win and Vanilla also indicates the importance of duplicating mechanisms.
    \item Compared with the pre-trained model, almost all the continuous learning methods demonstrate promising performance. This verifies the significant advantages of utilizing fresh samples for CVR prediction. Vanilla performs poorly on Criteo-30d but obtains a better result on Criteo-1d, possibly caused by the fact that the observed distribution deviated much more from the ground-truth distribution with longer $w_a$. This also indicates the importance of utilizing all conversions and performing unbiased estimation with delayed feedback modeling. Moreover, DEFER also demonstrates similar performance when compared with ES-DFM. Such a result can be attributed to the distribution bias caused by the long-term attribution window --- it ingests real negative samples from 30 days before into the training pipeline.
\end{itemize}

% \begin{itemize}[leftmargin=*]
%     \item \textbf{Impact of attribution window $w_a$}
%     \item \textbf{Independence modeling for Bi-DEFUSE}
% \end{itemize}

\begin{table}[t]
    \centering
    \caption{Performance of DEFUSE under different duplicating mechanisms in Criteo.}
    \begin{tabular}{l|r|r|r|r|}
         \toprule
        % \multirow{2}{*}{Methods}
        Dataset & \multicolumn{2}{c|}{Criteo-30d} & \multicolumn{2}{c|}{Criteo-1d} \\
        \midrule
         Metrics & AUC & RI-AUC & AUC & RI-AUC \\
         \midrule
         \midrule
         FNW & 0.8376 & 35.75\% & 0.8348 & 33.33\% \\
         FNW+DEFUSE & 0.8393 & 44.60\% & 0.8351 & 34.92\%  \\ %0.8388 &
         \midrule
         ES-DFM & 0.8396 & 46.11\% & 0.8459 & 92.06\% \\
         ES-DFM+DEFUSE & 0.8408 & 52.33\% & 0.8465 & 95.24\% \\
         \midrule
         DEFER &0.8382 & 38.86\% & 0.8463 & 94.17\% \\
         DEFER+DEFUSE & 0.8387 & 41.45\% & 0.8466 & 95.77\% \\
        %  \%Improv.  \\
         \bottomrule
    \end{tabular}
    \label{tab:rq3}
\end{table}

\subsection{Experiments under Different Duplicating Mechanisms (RQ2)}
Recall that, our DEFUSE is unbiased and can be applied to different duplicating mechanisms. To further verify the performance, we have conducted experiments using different training pipelines on Criteo dataset applied by FNW, ES-DFM and DEFER respectively, and report the AUC and RI-AUC results in Table~\ref{tab:rq3}. In general, by label correction and weighting the importance of four types of observed samples, our unbiased estimation demonstrates consistent improvement on performance among the three duplicating pipelines. 

\subsection{Study of DEFUSE (RQ3)}
% z_1、z_2的选择。wi mask是否有必要做？ 双头share vs 双头不共享；三种回补上应用DEFUSE后相比于FNX、ES-DFM、DEFER的改进；随着归因周期变化，双头DEFUSE的效果折线图分析。
Ablation studies on DEFUSE are also conducted to investigate the rationality and effectiveness of some designs --- to be more specific, (1) how different estimations of hidden variable $z$ may affect performance, (2) contributions of each components of Bi-DEFUSE, and (3) influence of different attribution window length $w_a$.

\begin{table}[t]
    \centering
    \caption{Impact of the ways of hidden variable inference $w.r.t.$ AUC and RI-AUC in Criteo.}
    \begin{tabular}{l|r|r|r|r|}
         \toprule
        % \multirow{2}{*}{Methods}
        Dataset & \multicolumn{2}{c|}{Criteo-30d} & \multicolumn{2}{c|}{Criteo-1d} \\
        \midrule
         Metrics & AUC & RI-AUC & AUC & RI-AUC \\
         \midrule
         \midrule
         DEFUSE \!+\! $z_{1}$ & 0.8408 & 52.33\% & 0.8465 & 96.30\% \\
         DEFUSE \!+\! $z_{2}$ & 0.8405 & 50.78\% & 0.8464 & 94.71\% \\
         DEFUSE \!+\! $z_{\mathrm{oracle}}$ & 0.8426 & 61.66\% & 0.8468 & 96.83\% \\
        %  \%Improv.  \\
         \bottomrule
    \end{tabular}
    \label{tab:ablation}
\end{table}

\subsubsection{\textbf{Impact of the estimation of $z(x)$}}
Since fake negatives are falsely labelled, a hidden variable $z$ is further introduced to infer whether the observed negatives are fake negatives. We hence implement an additional auxiliary model --- $z(x)$ that directly predicts FNs from observed negatives. As introduced in Section~\ref{sec:opt}, different choices of modelling $z(x)$ is experimented. Besides, to further explore the upper bound of our two-step optimization, we additionally 
investigate the ideal performance of DEFUSE given $z_{oracle} \in \{0, 1\}$, which indicates the ground-truth label for $z$. As shown in Table~\ref{tab:ablation}, here are some observations:
% z1的方差<z2
\begin{itemize}[leftmargin=*]
    \item DEFUSE$+z_1$ consistently outperforms DEFUSE+$z_2$. We credit such improvement to the reduction of high variance of $z(x)$. Prediction of $z_2(x)$ obviously involves divisions between two independent models, which may lead to unstable estimation and sub-optimal performance.
    \item DEFUSE$+z_{oracle}$ achieves best performance uniformly, this indicates the potential of optimizing our unbiased estimation by further improving the prediction of $z(x)$. 
    \item We also notice that the gap between $+z_{1}$ and $+z_{oracle}$ on Criteo-1d is smaller than that on Criteo-30d, indicating that lower proportion of fake negatives with relatively shorter attribution window makes it easier to estimate $z(x)$. 
\end{itemize}

\begin{figure}[tbp]
    \centering
    \includegraphics[width=.8\linewidth]{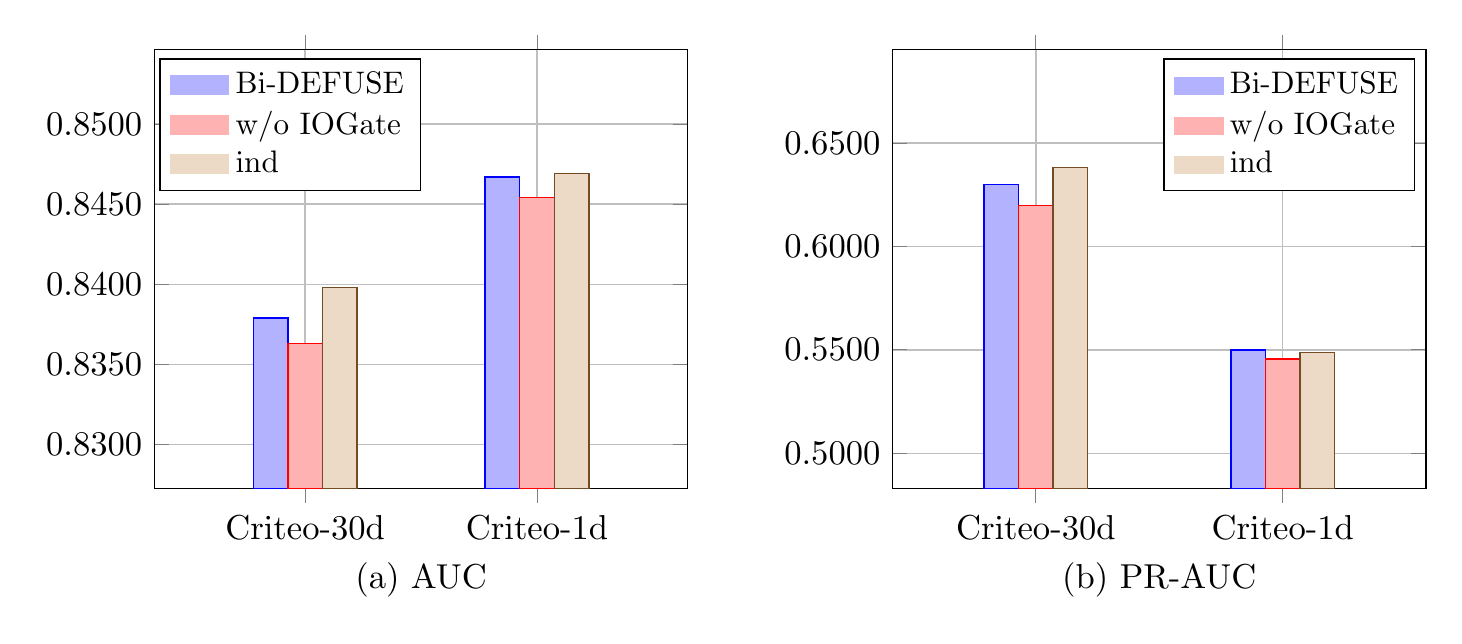}
    \caption{Ablation study for Bi-DEFUSE.}
    \label{fig:ablation}
\end{figure}

\subsubsection{\textbf{Contributions of different components in Bi-DEFUSE}}
\label{sec:aba}
To have a better understanding of Bi-DEFUSE, we do ablation study by evaluating three formulations of Bi-DEFUSE: (1)
Bi-DEFUSE as demonstrated in Figure~\ref{fig:bi-dis}; (2) a simpler version replacing the gates, experts and heads by MLPs; marked as $\mathrm{\text{w/o IOGate}}$; (3) completely disabling the shared network and estimating the in/out window CVR with two independent models, marked as $\mathrm{{\textit{ind}}}$. From Figure~\ref{fig:ablation}, we have:
%(3) we obtain another variant by $\ldots$. We summarize the experimental results in Figure~\ref{fig:ablation} and have the following findings:

\begin{itemize}[leftmargin=*]
    \item Removal of the MMoE gates and experts always degrades model performance since $\mathrm{Bi\!-\!DEFUSE}$ consistently outperforms $\mathrm{\text{w/o IOGate}}$.
    \item $\mathrm{{\textit{ind}}}$ uniformly dominates, indicating that IPs and DPs can be greatly different, independent models effectively avoid the conflict caused by predicting both IPs and long DPs using shared models. Yet, $\mathrm{{\textit{ind}}}$ is highly impractical in industrial scenarios, since it doubles the computational and storage consumption.
    \item Bi-DEFUSE still achieves comparable performance with $\mathrm{{\textit{ind}}}$ on Criteo-1d, likely a result of the fact that smaller $w_a$ not only greatly raises the ratio of IPs, making the unbiased estimation of IPs more decisive; but also effectively limits the difference between IPs and DPs, allowing the introduction of shared networks.
    %\item Although $\mathrm{{\text{ind}}}$ dominates, it is highly impractical in real industrial scenarios, since it doubles the computational and storage consumption. Note that Bi-DEFUSE still achieves comparable performance on Criteo-1d. One possible reason is the independent architecture can alleviate the biased caused by long-term attribution for out\_window modeling. More specifically, the in\_window modeling requires ground-truth distribution, but the out\_window modeling, which trained upon biased distribution, makes it rather difficult to train the shared networks.
    % \item 
\end{itemize}

\begin{figure}[tbp]
    \centering
    \includegraphics[width=.8\linewidth]{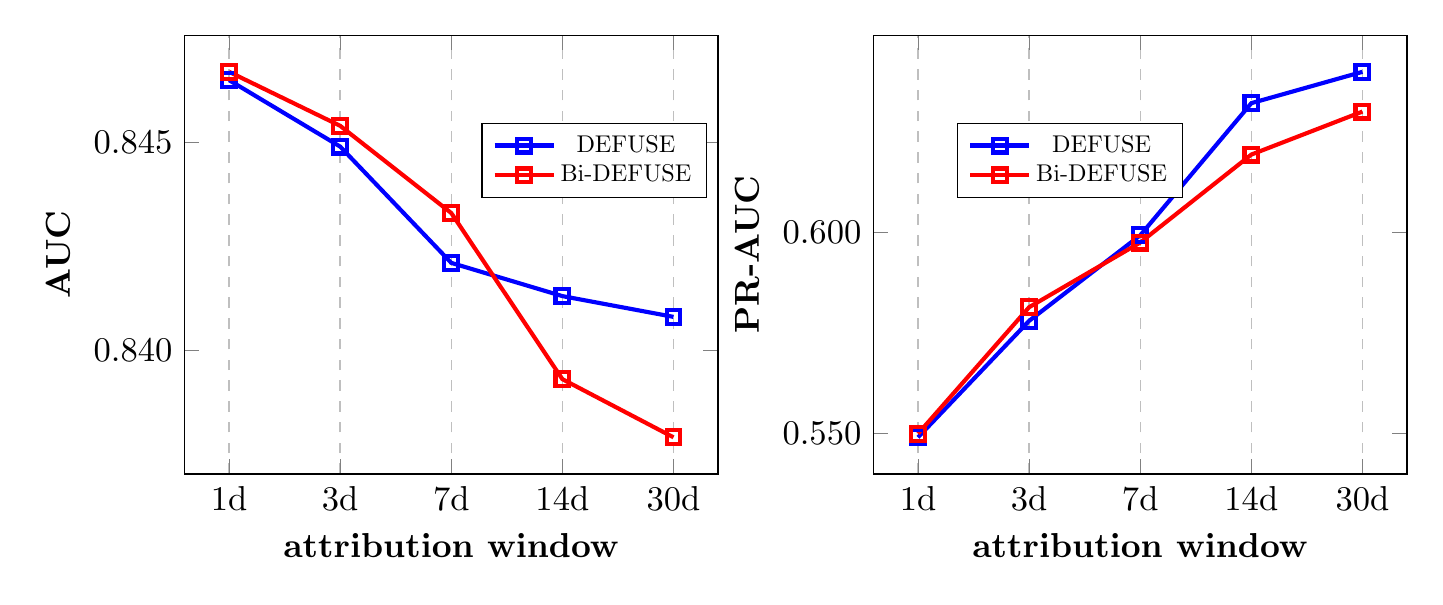}
    \caption{The effect of different window length for $w_a$.}
    \label{fig:atr}
\end{figure}

\subsubsection{\textbf{Performance of Bi-DEFUSE $w.r.t.$ different $w_a$.}}
As implied in Table~\ref{tab:perform}
%and  Figure~\ref{fig:ablation}
, performance of Bi-DEFUSE may degrade with long $w_a$. Towards this end, we investigate the influence of different attribution windows in depth. Since DEFUSE consistently outperforms baseline methods, for clear comparison, we only evaluate the performance of DEFUSE and Bi-DEFUSE with $w_a = \{1,3,7,14,30\}$ days on the Criteo dataset. Results reported in Figure~\ref{fig:atr} suggest that Bi-DEFUSE achieves the better performance with smaller $w_a$, e.g., $w_a \leq 7$ in terms of AUC and PR-AUC, since smaller $w_a$ not only makes the unbiased prediction of IPs more important, but also implies smaller challenges in estimating $z(x)$ and predicting DPs, which greatly amplifies the advantages of Bi-DEFUSE.

\subsection{Online Evaluation}
We conducted an A/B test in our online scenario to evaluate the proposed Bi-DEFUSE framework. We set 30min observation window and 1day attribution window and observed a steady performance improvement in terms of CVR(+2.28\%). This align with our offline streaming results and demonstrates the effectiveness of our DEFUSE in industrial systems.

% • Compared with KGAT-1w/o Att, KGAT-1w/o KGE performs better
% in most cases. One possible reason is that treating all neighbors
% equally (i.e., KGAT-1w/o Att) might introduce noises and mislead
% the embedding propagation process. It verifies the substantial
% influence of graph attention mechanism

% \begin{table}[t]
%     \centering
%     \caption{Impact of Independence $w.r.t.$ (RI-)AUC and (RI-)PR-AUC in Criteo 30d.}
%     \begin{tabular}{l|r|r|r|r}
%          \toprule
%         % \multirow{2}{*}{Methods}
%          Metrics & AUC & PR-AUC & RI-AUC & RI-PR-AUC \\
%          \midrule
%          \midrule
%          w/o ind & 0.8379 & 0.6301 \\
%          \midrule 
%          ind & 0.8398 & 0.6382 \\
%         %  \%Improv.  \\
%          \bottomrule
%     \end{tabular}
%     \label{tab:ablation}
% \end{table}

% 比较在co-graph和Inter-graph上学习structural information的效果
\section{Conclusion}
In this paper, we propose an asymptotically unbiased estimation method for the streaming CVR prediction, which addresses the delay feedback problem by weighting the importance of four types of observed samples. We carefully devise a new delayed feedback modeling approach, DEFUSE, with a two-step optimization mechanism. Specifically, we first infer the probability of fake negative  among observed negatives, then respectively correct the importance weights of each type of observed samples. Furthermore, to fully exploit the ground-truth immediate positives, we propose a bi-distribution modeling framework. Comprehensive experiments show that the proposed DEFUSE and Bi-DEFUSE significantly outperform the competitive baselines.

% \begin{acks}
% \end{acks}

% \newpage
\bibliographystyle{ACM-Reference-Format}
\bibliography{main}

\newpage
\appendix
\section{Appendix}

\subsection{Detailed demonstration of Equation~(\ref{eq:new_ub}) and derivations of $w$}

We first provide the demonstration of the equavilant of $\mathcal{L}_{ub}$, then show the detailed derivations for the importance weight $w_i(\mathbf{x},y)$, where $i \in \{IP,FN,RN,DP\}$.

\subsubsection{\textbf{Demonstration of $\mathcal{L}_{ub}$}}
Firstly, let $v$ denotes the observed label and $v_{IP}$, $v_{DP}$ respectively denote the label for immediate positive and delay positive sample, we formulate the minimization for $\mathcal{L}_{ub}$ as:
\begin{align}
\min_{\theta} \mathcal{L}_{ub} = & \min_{\theta} \! \int \! q(\mathbf{x}) \Bigl[ 
v_{IP}w_{IP}(\mathbf{x},y)\log f_{\theta}(\mathbf{x}) \notag \\ 
+& v_{IP}w_{IP}(\mathbf{x},y)\log p(d\leq w_o) \notag \\
+& (v_{DP}w_{DP}(\mathbf{x},y)+z(1-v)w_{FN}(\mathbf{x},y))\log f_{\theta}(\mathbf{x}) \notag \\
+& (v_{DP}w_{DP}(\mathbf{x},y)+z(1-v)w_{FN}(\mathbf{x},y))\log p(d > w_o) \notag \\
+& (1-v)(1-z)w_{RN}(\mathbf{x},y)\log(1-f_{\theta}(\mathbf{x}))
\Bigr] dx, 
\end{align}
where $z$ is the latent variable. Since the derivation of $\mathcal{L}_{ub}$ $w.r.t.$ $f_{\theta}$ is irrelevant with $\log p(d \leq w_o)$ and $\log p(d > w_o)$, it's trival to obtain following equivalence formula:
\begin{align}
& \min_{\theta} \mathcal{L}_{ub}  \notag \\
\Leftrightarrow & \min_{\theta} \! \int \! q(\mathbf{x}) \Bigl[
v_{IP}w_{IP}(\mathbf{x},y)\log f_{\theta}(\mathbf{x}) \notag \\
&+ (v_{DP}w_{DP}(\mathbf{x},y)+z(1-v)w_{FN}(\mathbf{x},y))\log f_{\theta}(\mathbf{x}) \notag \\
&+ (1-v)(1-z)w_{RN}(\mathbf{x},y)\log(1-f_{\theta}(\mathbf{x}))
\Bigr] dx \notag \\
= & \min_{\theta} \! \int \! q(\mathbf{x}) \Bigl[v(w_{DP}\log f_\theta(\mathbf{x}) + \mathbb{I}_{IP}(w_{IP}-w_{DP})\log f_\theta(\mathbf{x})) \notag\\
&+ (1-v)(w_{FN}\log f_\theta(\mathbf{x})z \notag\\
&+ w_{RN}\log (1-f_\theta(\mathbf{x}))(1-z)) \Bigr] dx,
\end{align}
Intuitively, to ensure an unbiased estimation, we must have $\mathcal{L}_{ub}$ from equation~(\ref{eq:ub2}) equals to $\mathcal{L}_{ideal}$ from equation~(\ref{eq:ideal}). That is,

\begin{align}
&\int \! q(\mathbf{x}) \Bigl[\sum_{v_i} q(v_i \! \mid \! x) w_i(\mathbf{x}, y(v_i, d)) \ell(\mathbf{x}, y(v_i,d); f_{\theta}(\mathbf{x})) \Bigr] dx \notag \\
=&
\int \! p(\mathbf{x}) \Bigl[\sum_{y(v_i,d)} p(y(v_i,d) \! \mid \! x)  \ell(\mathbf{x}, y(v_i,d); f_{\theta}(\mathbf{x})) \Bigr] dx,
\end{align}
Empirically, we can express the left part in the form of discrete summation as:
\begin{align}
\label{eq:left}
\mathcal{L}_{left} =& \sum \Bigl[ v_{IP}w_{IP}(\mathbf{x}, y)\log (f_{\theta}(\mathbf{x})p(d\leq w_o \mid \mathbf{x},y=1))  \notag \\
% +& v_{DP}w_{DP}(\mathbf{x}, y) + \log (f_{\theta}(\mathbf{x})p(d>w_o\mid \mathbf{x},y=1)) \notag \\
% +& z(\mathbf{x})(1-v)w_{FN}(\mathbf{x}, y)\log (f_{\theta}(\mathbf{x})p(d>w_o\mid \mathbf{x},y=1)) \notag \\
+& [v_{DP}w_{DP}(\mathbf{x}, y) + z(\mathbf{x})(1-v)w_{FN}(\mathbf{x}, y)] \ell_{DP}(\mathbf{x},y) \notag \\
%&+ y(1-v)w_{FN}(\mathbf{x}, y)\log (f_{\theta}(\mathbf{x})p(d>w_o\mid \mathbf{x},y=1)) \notag \\
+& (1-v)(1-z(\mathbf{x}))w_{RN}(\mathbf{x}, y)\log(1-f_{\theta}(\mathbf{x})) \Bigr].
\end{align}
where $\ell_{DP}(\mathbf{x},y) = \log (f_{\theta}(\mathbf{x})p(d>w_o \mid \mathbf{x},y=1))$ and $z(\mathbf{x})$ denotes fake negative probability. Similarly, we can express the right part as:
\begin{align}
\label{eq:right}
\mathcal{L}_{right} =& \sum \Bigl[ y_{IP}\log (f_{\theta}(\mathbf{x})p(d<w_o \mid \mathbf{x},y=1))  \notag \\
& + y_{DP}\log (f_{\theta}(\mathbf{x})p(d>w_o \mid \mathbf{x},y=1)) 
\notag \\
&+ (1-y)\log(1-f_{\theta}(\mathbf{x})) \Bigr]
%=& \sum_{q(\mathbf{x})} 
\end{align}

Take the middle term of equation~(\ref{eq:left},\ref{eq:right}) as an example, unbiased estimation would require either 
\begin{align}
\sum & [v_{DP}w_{DP}(\mathbf{x}, y) + z(\mathbf{x})(1-v)w_{FN}(\mathbf{x}, y)] \ell_{DP}(\mathbf{x},y) \notag \\
= \sum & y_{DP}\log (f_{\theta}(\mathbf{x})p(d>w_o\mid \mathbf{x},y=1)) \notag
\end{align} or
\begin{align}
\label{eq:forall}
v_{DP}(\mathbf{x})w_{DP}(\mathbf{x}, y)+z(\mathbf{x})(1-v)w_{FN}(\mathbf{x}, y)=y_{DP}(\mathbf{x}), \forall x.
\end{align}

Although directly solving $w$ from above equations, we can rewrite equation~(\ref{eq:forall}) into following expectation form:
\begin{align}
E_{q}[v_{DP}w_{DP}(\mathbf{x}, y)+z(1-v)w_{FN}(\mathbf{x}, y)]=E_{p}[y_{DP}].
\end{align} 

Emprically, equation always stands as well as the following unbiased estimations:

% Take derivative with respect to $f_{\theta}$ for specific $\mathbf{x}$ yields:

% Therefore, we can rewrite $\mathcal{L}_{left}$ as:
% \begin{align}
% \mathcal{L}_{left} =& \sum \Bigl[ v_{IP}w_{IP}(\mathbf{x}, y)\log (f_{\theta}(\mathbf{x}))  \notag \\
% +& v_{DP}w_{DP}(\mathbf{x}, y) + \log (f_{\theta}(\mathbf{x})) \notag \\
% +& z(\mathbf{x})(1-v)w_{FN}(\mathbf{x}, y)\log (f_{\theta}(\mathbf{x})) \notag \\
% % & + [v_{DP}w_{DP}(\mathbf{x}, y)+z(\mathbf{x})(1-v)w_{FN}(\mathbf{x}, y)]\log (f_{\theta}(\mathbf{x})p(d>w_o\mid \mathbf{x},y=1))
% % \notag \\
% %&+ y(1-v)w_{FN}(\mathbf{x}, y)\log (f_{\theta}(\mathbf{x})p(d>w_o\mid \mathbf{x},y=1)) \notag \\
% +& (1-y)(1-v)w_{RN}(\mathbf{x}, y)\log(1-f_{\theta}(\mathbf{x})) \Bigr].
% \end{align}

% \begin{align}
% &\sum y w_{TP} \log f_{\theta}(\mathbf{x}) \notag \\
% &+ \sum (1-y) \left[z_i w_{FN} \log (f_{\theta}(\mathbf{x}) p(d>w_o)) + (1-z_i) w_{RN} \log (1-f_{\theta}(\mathbf{x})\right] \\
% =& \sum y v w_{P} \log (f_{\theta}(\mathbf{x})p(d \leq w_o)) \notag \\
% &+ \sum \left[ y(1-v)w_{DP} + (1-y)z_i w_{FN} \right] \log (f_{\theta}(\mathbf{x}) p(d>w_o)) \notag \\
% &+ \sum (1-y)(1-z_i) w_{RN} \log (1-f_{\theta}(\mathbf{x}))
% \end{align}

\begin{align}
\mathbb{E}_{q}\left[v_{DP} \right](\mathbf{x}) &= \frac{f_{dp}(\mathbf{x})}{1+f_{dp}(\mathbf{x})} \\
\mathbb{E}_{q}\left[z \right](\mathbf{x}) &= \frac{f_{dp}(\mathbf{x})}{1-p_{win}(\mathbf{x})}\\
\mathbb{E}_{p}\left[y_{DP} \right](\mathbf{x}) &= f_{dp}(\mathbf{x}) \\
\mathbb{E}_{q}\left[1-v \right](\mathbf{x}) &= \mathbb{E}_{q}\left[1-(v_{IP} + v_{DP}) \right](\mathbf{x}) = \frac{1-p_{win}(\mathbf{x})}{1 + f_{dp}(\mathbf{x})},
\end{align}
where $p_{win}(\mathbf{x}) = p(y=1 \mid \mathbf{x}) p(d < w_o\mid \mathbf{x},y=1)$. With sufficient data, which is usually abundant in digital advertising, we may reasonably assume following asymptotically unbiased estimation:
\begin{align}
 \frac{f_{dp}(\mathbf{x})}{1+f_{dp}(\mathbf{x})}w_{DP}(\mathbf{x}, y)+ \frac{f_{dp}}{1-p_{win}}\frac{1-p_{win}}{1 + f_{dp}(\mathbf{x})}w_{FN}(\mathbf{x}, y)\longrightarrow f_{dp}(\mathbf{x}).
\end{align}

This leads to $w_{DP}(\mathbf{x}, y)+w_{FN}(\mathbf{x}, y)=1+f_{dp}(\mathbf{x})$. Likewise, we may derive that $w_{IP}(\mathbf{x},y)=w_{RN}(\mathbf{x},y)=1+f_{dp}(\mathbf{x})$.

\subsection{Unbiased Estimation for other duplicating mechanisms}
Next, we state the unbiased estimation for other duplicating mechanisms(FNW+DEFUSE, DEFER+DEFUSE). Since the derivation process is similar, we directly show the loss form.

\subsubsection{$\mathcal{L}_{ub}$ for dup-mechanism in FNW}
For unbiased estimation for the dup-mechanism in FNW, since the size of $w_o$ is 0, we divide the observed samples into $DP$, $FN$, and $RN$. We have following unbiased loss form:
\begin{align}
& \min_{\theta} \mathcal{L}_{ub}  \notag \\
\Leftrightarrow & \min_{\theta} \! \int \! q(\mathbf{x}) \Bigl[
vw_{DP}(\mathbf{x},y)\log f_{\theta}(\mathbf{x}) \notag \\
&+ z(1-v)w_{FN}(\mathbf{x},y))\log f_{\theta}(\mathbf{x}) \notag \\
&+ (1-v)(1-z)w_{RN}(\mathbf{x},y)\log(1-f_{\theta}(\mathbf{x}))
\Bigr] dx,
\end{align}
$s.t.$ 
\begin{align*}
    w_{RN}(\mathbf{x},y) &= 1 + f_{dp-fnw}(\mathbf{x}) \\
    w_{DP}(\mathbf{x},y) + w_{FN}(\mathbf{x},y) &= 1 + f_{dp-fnw}(\mathbf{x}),
\end{align*}
where $f_{dp-fnw} = f_{\theta}$ denotes the probability of observed negative samples to be fake negative. Emprically, we set $w_{DP}(\mathbf{x},y) = 1$ since DP can be observed, and reduce the importance weight of fake negative($w_{FN}(\mathbf{x},y)$) to $f_{\theta}$.

\subsubsection{$\mathcal{L}_{ub}$ for dup-mechanism in DEFER}
The loss form of unbiased DEFER loss is same to ES-DFM+DEFUSE, as they both have observation window $w_o$ and attribution window $w_a$. The main difference lies in the way of duplication for real negative samples, which will lead to different estimation forms of importance weights.
\begin{align}
& \min_{\theta} \mathcal{L}_{ub}  \notag \\
\Leftrightarrow & \min_{\theta} \! \int \! q(\mathbf{x}) \Bigl[
v_{IP}w_{IP}(\mathbf{x},y)\log f_{\theta}(\mathbf{x}) \notag \\
&+ (v_{DP}w_{DP}(\mathbf{x},y)+z(1-v)w_{FN}(\mathbf{x},y))\log f_{\theta}(\mathbf{x}) \notag \\
&+ (1-v)(1-z)w_{RN}(\mathbf{x},y)\log(1-f_{\theta}(\mathbf{x}))
\Bigr] dx,
\end{align}
$s.t.$ 
\begin{align*}
    w_{IP}(\mathbf{x},y) = w_{RN}(\mathbf{x}) &= 2 \\
    w_{DP}(\mathbf{x},y) + w_{FN}(\mathbf{x}) &= 2.
\end{align*}
The weights of both IPs and RNs are duplicated. Moreover, since DP samples has same feature distribution with FN samples, we emprically set $w_{DP}(\mathbf{x},y) = w_{FN}(\mathbf{x},y) = 1$.

\end{document}